\documentclass{article}

\usepackage[final]{neurips_2021}

\usepackage[utf8]{inputenc} % allow utf-8 input
\usepackage[T1]{fontenc}    % use 8-bit T1 fonts
\usepackage{hyperref}       % hyperlinks
\usepackage{url}            % simple URL typesetting
\usepackage{booktabs}       % professional-quality tables
\usepackage{amsfonts}       % blackboard math symbols
\usepackage{nicefrac}       % compact symbols for 1/2, etc.
\usepackage{microtype}      % microtypography

\usepackage{float}
\usepackage{upgreek}

\usepackage{setspace}
\usepackage{amsmath}
\usepackage{amssymb} 
\usepackage{amsthm}
\usepackage{amsfonts, mathtools}
\usepackage{algorithm,algpseudocode}
\usepackage{bm}
\usepackage{graphicx}
\usepackage{multirow, booktabs}
\usepackage{caption}
\usepackage{subcaption}
\usepackage[colorinlistoftodos]{todonotes}
\usepackage{standalone}

\DeclareMathOperator*{\argmin}{arg\,min}

\newtheorem{proposition}{Proposition}
\newtheorem{lemma}{Lemma}

\newtheorem{condition}{Condition}

\newtheorem{theorem}{Theorem}

\title{An Adaptive State Aggregation Algorithm for Markov Decision Processes}

\author{Guanting Chen$^\dagger$ \ \ \ \   Johann Demetrio Gaebler$^\dagger$\ \ \ \  Matt Peng$^\ddagger$\ \ \ \  Chunlin Sun$^\dagger$ \ \  \ \ Yinyu Ye$^\S$ \\
$^\dagger$ Institute for Computational and Mathematical Engineering, Stanford University\\
$^\ddagger$ Department of Electrical Engineering \& Computer Sciences, University of California, Berkeley\\
$^\S$ Department of Management Science and Engineering, Stanford University \\
$\{$guanting, jgaeb, chunlin, yyye$\}$@stanford.edu\,\,\, mattpeng@berkeley.edu\\
}

\begin{document}

\maketitle

\begin{abstract}
Value iteration is a well-known method of solving Markov Decision Processes (MDPs) that is simple to implement and boasts strong theoretical convergence guarantees. However, the computational cost of value iteration quickly becomes infeasible as the size of the state space increases. Various methods have been proposed to overcome this issue for value iteration in large state and action space MDPs,
often at the price, however, of generalizability and algorithmic simplicity.
In this paper, we propose an intuitive algorithm for solving MDPs that reduces the cost of value iteration updates by dynamically grouping together states with similar cost-to-go values.
We also prove that our algorithm converges almost surely to within \(2\varepsilon / (1 - \gamma)\) of the true optimal value in the \(\ell^\infty\) norm, where \(\gamma\) is the discount factor and aggregated states differ by at most \(\varepsilon\).
Numerical experiments on a variety of simulated environments confirm the robustness of our algorithm and its ability to solve MDPs with much cheaper updates especially as the scale of the MDP problem increases.
\end{abstract}

\section{Introduction}
\label{sec:intro}

State aggregation is a long-standard approach to solving large-scale
% systems.
Markov decision processes (MDPs).
The main idea of state aggregation is to define the similarity between states, and work with a system of reduced complexity size by grouping similar states into aggregate, or ``mega-'', states.
Although there has been a variety of results on the performance of the policy using state aggregation \cite{li2006, van2006, abel2016}, a common assumption is
%a pre-specified aggregation scheme.
that states are aggregated according to the similarity of their optimal cost-to-go values. Such a scheme, which we term \emph{pre-specified aggregation}, is generally infeasible unless the MDP is already solved. %Solving MDPs especially as they grow much larger requires not only lots of memory but also lots of compute resources would negatively impact the environment.
%Solving MDPs more efficiently then leads to uniformly better algorithms.
To address the difficulties inherent in pre-specified aggregation, algorithms that learn how to effectively aggregate states have been proposed \cite{ortner2013, duan2018, sinclair2019}.

This paper aims to provide new insights into
adaptive
online learning of the correct state aggregates.
We propose a simple and efficient state aggregation algorithm for calculating the optimal value and policy of an infinite-horizon discounted MDP that can be applied in planning problems \cite{baras2000} and generative MDP problems \cite{sidford2018near}.
The algorithm alternates between two distinct phases.
During the first phase (``global updates''), the algorithm
updates the cost-to-go values as in the standard value iteration,
trading off some efficiency to more accurately guide the cost-to-go values in the right direction;
in the second phase (``aggregate updates''), the algorithm groups together
states with similar cost-to-go values based on the last sequence of global updates then efficiently updates the states in each mega-state in tandem as it optimizes over the reduced space of aggregate states.

The algorithm is simple: the inputs needed for state aggregation are the current cost-to-go values and the parameter $\varepsilon$, which bounds the difference between (current) cost to go values of states within a given mega-state.
Compared to prior works on state aggregation that use information on the state-action value ($Q$-value) \cite{sinclair2019}, transition density \cite{ortner2013}, and methods such as upper confidence intervals, our method does not require strong assumptions or extra information, and only performs updates in a manner similar to standard value iteration.

Our contribution is a feasible online algorithm for learning aggregate states and cost-to-go values that requires no extra information beyond that required for standard value iteration.
We showcase in our experimental results that our method provides significantly faster convergence than standard value iteration especially for problems with larger state and action spaces. We also
provide theoretical guarantees for the convergence, accuracy, and convergence rate of our algorithm.

The simplicity and robustness of our novel state aggregation algorithm demonstrates its utility and general applicability in comparison to existing approaches for solving large MDPs.
%We believe, to our knowledge, that our contribution in a feasible and practical randomized SA algorithm will lay the path forward for future works along this direction that can build off our foundation.

\subsection{Related literature}
\label{sec:lit}

When a pre-spcified aggregation is given, \cite{tsitsiklis1996, van2006} give performance bounds on the aggregated system and propose variants of value iteration.
There are also a variety of ways to perform state aggregation based on different criteria. \cite{ferns2012, dean1997} analyze partitioning the state space by grouping states whose transition model and reward function are close. \cite{mccallum1997} proposes aggregate states that have the same optimal action
and similar $Q$-values for these actions. \cite{jong2005} develops aggregation techniques such that states are aggregated if they have the
same optimal action.
We refer the readers to \cite{li2006, abel2016, abel2020} for a more comprehensive survey on this topic.

On the dynamical learning side, our adaptive state-aggregated value iteration algorithm is also related to the
%aggregating-disaggregation
aggregation-disaggregation method used to accelerate the convergence of value iterations (\cite{bertsekas1988,schweitzer1985,mendelssohn1982,bertsekas2018}) in policy evaluation, i.e., to evaluate the value function of a policy.
Among those works,
%the work
that of \cite{bertsekas1988} is closest to our approach.
Assuming the underlying Markov processes to be ergodic, the authors propose to group states based on Bellman residuals in between runs of value iteration.
They also allow states to be aggregated or disaggregated at every abstraction step.
However, the value iteration step in our setting is different from the value iteration in \cite{bertsekas1988} for the reason that value iteration in policy evaluation will not involve the Bellman operator.
Aside from state aggregation, a variety of other methods have been studied to accelerate value iteration. \cite{herzberg1994} proposes
iterative algorithms based on a one-step look-ahead approach. \cite{shlakhter2010} combines the so called ``projective operator'' with value iteration and achieve better efficiency.
\cite{anderson1965,fang2009,zhang2020} analyze the Anderson mixing approach to speed up the convergence of fixed-point problems.

Dynamic learning of the %aggregation
aggregate states has also been studied %in the more
more generally in MDP and reinforcement learning (RL) settings. \cite{hostetler2014} proposes a class of \(Q\)-value--based state aggregations and applies them to Monte Carlo tree search. \cite{slivkins2011} uses data-driven
discretization to adaptively discretize state and action space in a contextual bandit setting. \cite{ortner2013} develops an algorithm for learning state aggregation in an online setting by leveraging confidence intervals. \cite{sinclair2019} designs a \(Q\)-learning algorithm based on data-driven adaptive discretization of the state-action space.
For more state abstraction techniques see \cite{baras2000,dean1997,jiang2014,abel2019}.

% Based on policy iteration, Bertsekas developed an adaptive aggregation scheme for policy evaluation.
% i.e., given a fixed policy, using state aggregation to find the corresponding cost-to-go. However, its selection
% of the aggregation matrix is based on the assumption of ergodic processes and its number of aggregated
% groups is fixed. Furthermore, no aggregation is applied to the policy improvement step, which may limit the
% speed of the algorithm

\section{Preliminaries}
\label{sec:prelim}

\subsection{Markov decision process}
We consider an infinite-horizon Markov Decision Process $M = (\mathcal{S}, \mathcal{A}, P, r,\gamma,\rho)$, consisting of: a finite state space $\mathcal{S}$; a finite action space $\mathcal{A}$; the probability transition model $P$, where $P(s'|s,a)$ denotes the probability of transitioning to state $s'$ conditioned on the state-action pair $(s, a)$; the immediate cost function $r : \mathcal{S} \times \mathcal{A} \to \mathbb{R}$, which denotes the immediate cost---or reward---obtained from a particular state-action pair; the discount factor $\gamma \in [0,1)$; and the initial distribution over states in $\mathcal{S}$, which we denote by $\rho$.

A policy $\pi : \mathcal{S} \to \mathcal{A}$ specifies the agent's action based on the current state,
%and the policy could be deerministic or stochastic.
either deterministically or stochastically.
A policy induces a distribution over the trajectories $\tau = (s_t, a_t, r_t)_{t=0}^{\infty}$, where $s_0 \sim \rho$, and $a_t \sim \pi(\cdot | s_t)$ and $s_{t+1} \sim P(\cdot|s_t, a_t)$ for $t \geq 0$.

A value function $\bm{V}:\mathcal{S}\to\mathbb{R}^{|\mathcal{S}|}$ assigns a value to each state;
as \(|\mathcal{S}| < \infty\), $\bm{V}$ can also be represented by a finite-length vector $(V(1),...,V(|\mathcal{S}|))^{\top}\in\mathbb{R}^{|\mathcal{S}|}$.
(In this paper, we view all vector as vector functions mapping from the index to the corresponding entry.) Moreover, each policy $\pi$ is associated with a value function $\bm{V}^{\pi} : \mathcal{S} \to \mathbb{R}$, which is defined to be the discounted sum of future rewards starting at $s$ and with policy $\pi$:
$$\bm{V}^{\pi}(s) := \mathbb{E}\left[\sum_{t=0}^{\infty}\gamma^tr(s_t,a_t)|\pi,s_0 = s\right].$$  
As noted above, we represent both the value function corresponding to the policy $\pi$ and the value vector $(V^{\pi}(1),...,V^{\pi}(|\mathcal{S}|))^{\top}$ as $\bm{V}^{\pi}$.

%For each action $a\in\mathcal{A}$ and any state $s\in\mathcal{S}$,
For each state \(s \in \mathcal{S}\) and action \(a \in \mathcal{A}\) belonging to state \(s\),
let $\bm{P}_{s,a} \in \mathbb{R}^{|\mathcal{S}|}$ denote the vector of transition probabilities resulting from taking the action $a$ in state $s$.
We call a policy $\pi$ \emph{greedy} with respect to a given value $\bm{V}\in\mathbb{R}^{|\mathcal{S}|}$ if 
$$\pi(s) \in \argmin_{a\in\mathcal{A}}\left(r(s, a) + \gamma \cdot \bm{P}_{s,a}^{\top}\bm{V})\right).$$
We define $\bm{T}: \mathbb{R}^{|\mathcal{S}|} \to \mathbb{R}^{|\mathcal{S}|}$ to be the dynamic programming operator such that $(\bm{T}\bm{V})(s) = T_s(\bm{V})$ where 
$$T_s(\bm{V}) = \min_{a\in \mathcal{A}}\left(r(s, a) + \gamma \cdot \bm{P}_{s,a}^{\top} \bm{V}\right).$$

% We also define the $Q$-value function $Q^{\pi} : \mathcal{S} \times \mathcal{A} \to \mathbb{R}$ as
% $$Q^{\pi}(s, a) = \mathbb{E}\left[\sum_{t=0}^{\infty} \gamma^tr(s_t, a_t)|\pi,s_0 = s, a_0 = a\right],$$
% where $a_t = \pi(s_t)$ for $t>0$.

The optimal value function $\bm{V}^*$ is the unique solution of the equation $\bm{V}^* = T\bm{V}^*$. A common approach to find $\bm{V}^*$ is value iteration, which, given an initial guess $\bm{V}_0$, generates a sequence of value functions $\{\bm{V}_t\}_{t=0}^\infty$ such that $\bm{V}_{t+1} = \bm{T}\bm{V}_t$.
The sequence $\{\bm{V}_t\}_{t=0}^{\infty}$ converges to $V^*$ as $t$ goes to $+\infty$ \cite{Bel}.

\subsection{State aggregation}

The state space of MDPs
%are typically
can be very large.
State aggregation divides the state space $\mathcal{S}$ into $K$ subsets and views each collection of states as a mega-state.
Then, the value function generated by the mega-states can be used to approximate the optimal value $\bm{V}^*$.

To represent a state aggregation, we define the matrix $\bm{\Phi} \in \mathbb{R}^{|\mathcal{S}|\times K}$.
%The $i$-th column consists of $\{1, 0\}$ values indicating whether or not each state is in the $i$-th mega-state.
We set \(\phi_{i,j} = 1\) if state \(i\) is in the \(j\)-th mega-state, and let \(\phi_{i,j} = 0\) otherwise; i.e., column \(j\) of \(\bm{\Phi}\) indicates whether each state belongs to mega-state \(j\).
The state-reduction matrix $\bm{\Phi}$ also induces a partition $\{S_i\}_{i=1}^K$ on \(\mathcal{S}\), i.e., $\mathcal{S} = \bigcup_{i=1}^K S_i$ and $S_i \cap S_j = \emptyset$ for $i \neq j$. Denote by $\bm{W} \in \mathbb{R}^K$ the cost-to-go value function for the aggregated state, and note that the current value of $\bm{W}$ induces a
%n approximated
value function $\tilde{\bm{V}}(\bm{W}) \in \mathbb{R}^{|\mathcal{S}|}$ on the original state space, where
$$\tilde{\bm{V}}(s, \bm{W}) = \bm{W}(j), \,\, \text{ for }s\in S_j.$$

\section{Algorithm design}
\label{sec:sa}

In this section, we first introduce a state aggregation algorithm
%knowing
which assumes knowledge of the optimal value function. The algorithm is proposed in \cite{tsitsiklis1996} which also provides the corresponding convergence result.
Based on existing theory, we design our adaptive algorithm and discuss its convergence properties.

\subsection{A pre-speficied aggregation algorithm}
Given a pre-specified aggregation, we seek the value function $\bm{W}$ for the aggregated states such that 
\begin{align}\label{eqn_prefix_error}
    \|\tilde{\bm{V}}(\bm{W}) - \bm{V}^*\|_{\infty} = O(\|\bm{e}\|_{\infty}),
\end{align}
where $\bm{e}=(e_1,...,e_K)^{\top}$ and $e_j = \max_{s_1, s_2 \in S_j}|V^*(s_1) - V^*(s_2)|$ for $j=1,...,K$. Intuitively, Eq.~$\eqref{eqn_prefix_error}$ justifies our approach of %state aggregation: to aggregate
aggregating states that have similar optimal cost-to-go values.
We then state the algorithm that will converge to the correct cost-to-go values while satisfying Eq.~$\eqref{eqn_prefix_error}$.

\begin{algorithm}[ht!]
\caption{Random Value Iteration with Aggregation}
\label{alg:RVIA}
\begin{algorithmic}[1]
\State Input: $\bm{P}$, $r$, $\gamma$, $\bm{\Phi}$, $\{\alpha_t\}_{t=1}^{\infty}$
\State Initialize $\bm{W}_0 = \bm{0}$ 
\For {$t=1,..., n$}
\For {$j=1,..., K$}
\State Sample state $s$ uniformly from set $S_j$
\State $W_{t+1}(j) = (1-\alpha_t)W_{t}(j) + \alpha_t T_s\tilde{\bm{V}}(\bm{W}_{t})$
\EndFor
\EndFor
\State Output: $\tilde{\bm{V}}_n$
\end{algorithmic}
\end{algorithm}

Algorithm \ref{alg:RVIA} takes a similar form in Stochastic Approximation \cite{robbins1951,wasan2004}, and will converge almost surely to a unique cost-to-go value. Here $\alpha_t$ is the step size of the learning algorithm; by taking, e.g., $\alpha_t = 1$, we recover the formula of value iteration. %The convergence %of the algorithm
The following convergence result is proved in \cite{tsitsiklis1996}.
%, and we state the result below.

\begin{proposition}[Theorem 1, \cite{tsitsiklis1996}]
\label{prop_prefix}
When $\sum_{t=1}^{\infty} \alpha_t = \infty$ and $\sum_{t=1}^{\infty} \alpha_t^2 < \infty$, $\{\bm{W}_t\}_{t=1}^{\infty}$ in Line~6 of Algorithm~\ref{alg:RVIA}  will converge almost surely to $\bm{W}^*$ entry-wise, where \(\bm{W}^*\) is the solution of 
\begin{equation}
    W^*(j) = \cfrac{1}{|S_j|}\sum_{s\in S_j}T_s\tilde{\bm{V}}(\bm{W}^*).
\end{equation}
Define $\pi^{\bm{W}^*}$ to be the greedy policy with respect to $\tilde{\bm{V}}(\bm{W}^*)$. Then, we have, morevoer, that
\begin{equation}\label{ineq_error}
    \begin{aligned}
    \|\tilde{\bm{V}}(\bm{W}^*) - \bm{V}^*\|_{\infty} &\leq \frac{\|\bm{e}\|_{\infty}}{1-\gamma},\\
    \|{\bm{V}}^{\pi^{\bm{W}^*}} - \bm{V}^*\|_{\infty} &\leq \frac{2\gamma\|\bm{e}\|_{\infty}}{(1-\gamma)^2},
    \end{aligned}
\end{equation}
where ${\bm{V}}^{\pi^{\bm{W}^*}}$ is the value function associated with policy $\pi^{\bm{W}^*}$.
\end{proposition}

Proposition 1 states that if we are able to partition the state space such that the maximum difference of the optimal value function within each mega-state is small, the %approximated
value function produced by Algorithm~\ref{alg:RVIA} can approximate the optimal value up to $\frac{\|\bm{e}\|_{\infty}}{1-\gamma}$, and the policy associated with the approximated value function will also be close to the optimal policy.

\subsection{Value iteration with state aggregation}

In order to generate an efficient approximation, Proposition \ref{prop_prefix} requires a pre-specified aggregation scheme such that $\max_{s_1, s_2 \in S_i}|V^*(s_1) - V^*(s_2)|$ is small for every $i$ to guarantee the appropriate level of convergence for Algorithm~\ref{alg:RVIA}. 
Without knowing $\bm{V}^*$, is it still possible to control the approximation error?
In this section we answer in the affirmative by introducing an adaptive state aggregation scheme that learns the correct state aggregations online as it learns the true cost-to-go values.

Given the current cost-to-go value vector $V \in \mathbb{R}^{|\mathcal{S}|}$, let $b_1 = \min_{s \in \mathcal{S}} V(s)$, let $b_2 = \max_{s \in \mathcal{S}} V(s)$. Group the cost-to-go values among disjoint subintervals of the form $\lceil(b_2 - b_1)/\varepsilon\rceil$.
Next, let $\Delta = (b_2 - b_1)/\varepsilon$, and let $S_j$ to be the $j$-th mega-state, which contains all the states whose current estimated cost-to-go value falls in the interval $[b_1 + (j-1)\varepsilon, b_1 + j\varepsilon)$.
Grouping the states in this way reduces the state size from $|\mathcal{S}|$ states to at most $\lceil(b_2 - b_1)/\varepsilon\rceil$ mega-states. See Algorithm~\ref{alg:A} for further details.

\begin{algorithm}[ht!]
\caption{Value-based Aggregation}
\label{alg:A}
\begin{algorithmic}[1]
\State Input: $\varepsilon$, $\bm{V}=(V(1),...,V(|\mathcal{S}|))^{\top}$
\State $b_1 = \min\limits_{s \in |\mathcal{S}|} V(s)$, $b_2 = \max\limits_{s \in |\mathcal{S}|} V(s)$, $\Delta = (b_2 - b_1)/\varepsilon$
\For {$i=1,..., \lceil\Delta\rceil$}
\State $\hat{S}_i = \left\{s | V(s) \in [b_1 + (i-1)\varepsilon, b_1 + i\varepsilon)\right\}$, $\hat{W}(i) = b_1 + (i-\frac{1}{2})\varepsilon$
\EndFor
\State Delete the empty sets in $\{\hat{S}_i\}_{i=1}^{\lceil\Delta\rceil}$ while keep the same order, and define the modified partition to be $\{S_i\}_{i=1}^{K}$, where $K$ is the cardinally of the modified set of mega-states. Modify $\hat{W}$ and generate $W \in \mathbb{R}^K$ the similar way.
\State Return $\{S_i\}_{i=1}^{K}$ and $W$.
\end{algorithmic}
\end{algorithm}

Without the knowledge of $\bm{V}^*$ in advance, one must periodically perform value iteration on $\mathcal{S}$ to learn the correct aggregation to help with adapting the aggregation scheme. As a result, our algorithm alternates between two phases: in the global update phase the algorithm performs value iteration on $\mathcal{S}$; in the aggregated update phase, the algorithm starts to group together states with similar cost-to-go values based on the result of the last global update, and then performs aggregated updates as in Algorithm~\ref{alg:RVIA}. 

We denote by $\{\mathcal{A}_i\}_{i=1}^{\infty}$ the intervals of iterations in which the algorithm performs state-aggregated updates, and we denote by $\{\mathcal{B}_i\}_{i=1}^{\infty}$ the intervals of iterations in which the algorithm performs global update.
As a consequence, $b < a$ for any $a\in\mathcal{A}_i$ and $b\in\mathcal{B}_i$; likewise, $a < b$ for any $a\in\mathcal{A}_i$ and $b\in\mathcal{B}_{i+1}$.

We then present our adaptive algorithm. For a pre-speficied number of iterations $n$, the time horizon $[1, n)$ is divided into intervals of the form $\mathcal{B}_1, \mathcal{A}_1, \mathcal{B}_2, \mathcal{A}_2,\ldots$. Every time the algorithm exits an interval of global updates $\mathcal{B}_i$, it runs Algorithm~\ref{alg:A} based on the current cost-to-go value and the parameter $\varepsilon$, using the output of Algorithm~\ref{alg:A} for the current state aggregation and cost-to-go values for $\mathcal{A}_i$. Similarly, every time the algorithm exits an interval of state-aggregated updates $\mathcal{A}_i$, it sets $\tilde{\bm{V}}(\bm{W})$, where $\bm{W}$ is the current cost-to-go value for the aggregated space, as the initial cost-to-go value for the subsequent interval of global iterations.

\begin{algorithm}[ht!]
\caption{Value Iteration with Adaptive Aggregation}
\label{alg:AVIA}
\begin{algorithmic}[1]
\State Input: $P$, $r$, $\varepsilon$, $\gamma$, $\{\alpha_t\}_{t=1}^{\infty}$, $\{\mathcal{A}_i\}_{i=1}^{\infty}$, $\{\mathcal{B}_i\}_{i=1}^{\infty}$
\State Initialize $W_0 = \bm{0}$, $V_1 = \bm{0}$, $t_{sa} = 1$
\For {$t=1,..., n$}
    \If {$t \in \mathcal{B}_i$}
    \If {$t = \min\{\mathcal{B}_i\}$} $\bm{V}_{t-1} = \tilde{\bm{V}}(\bm{W}_{t-1}).$
    \EndIf
    \For {$j=1,..., |\mathcal{S}|$}
    $V_t(j) = T_j\bm{V}_{t-1}.$
    \EndFor
    \Else
    \State Find current $i$ s.t. $t \in \mathcal{A}_i$
    \If {$t = \min\{\mathcal{A}_i\}$} 
    \State Define $\{S_i\}_{i=1}^{K}$ and $\bm{W}_t$ to be the output of Algorithm \ref{alg:A} with input $\varepsilon$, $\bm{V}_{t-1}$.
    \EndIf
    \For {$j=1,..., K$}
    \State Sample state $s$ uniformly from set $S_j$.
    \State \begin{align}\label{eqn_W_adaptive}
        W_t(j) = (1-\alpha_{t_{sa}}){W}_{t-1}(j) + \alpha_{t
        _{sa}} T_s\tilde{\bm{V}}(\bm{W}_{t-1})
    \end{align}
    \EndFor
    \State $t_{sa} = t_{sa} + 1$
    \EndIf
\EndFor
\If {$n \in \mathcal{B}_i$} return $\bm{V}_n$.
\EndIf
\State return $\tilde{\bm{V}}(\bm{W}_n)$.
\end{algorithmic}
\end{algorithm}

\subsection{Convergence}
From Proposition \ref{prop_prefix}, if we fix the state-aggregation parameter $\varepsilon$, even with perfect information, state aggregated value iteration will generate an approximation of the cost-to-go values with \(\ell^\infty\) error bounded by $\varepsilon/(1-\gamma)$.
This bound is sharp, as shown in \cite{tsitsiklis1996}.
%make an example that the error bound $\|\bm{e}\|_{\infty}/(1-\gamma)$ holds with equality.
Such error is negligible in the early phase of the algorithm, but the error would accumulate in the later phase of the algorithm and prevent the algorithm from converging to the optimal value. As a result, it is not desirable for $\limsup |\mathcal{A}_t| \to \infty$.\footnote{
    By setting $\varepsilon$ adaptively, one might achieve better complexity and error bound by setting  $\lim\sup |\mathcal{A}_t| \to \infty$;
    however, adaptively choosing \(\varepsilon\) lies beyond the scope of this work.
}

We state asymptotic convergence results for Algorithm \ref{alg:AVIA}; proofs can be found in the supplementary materials.
For the remainder of the paper, by a slight abuse of notation, we denote by $\bm{V}_t$ the current cost-to-go value at iteration $t$.
More specifically, if the current algorithm is in phase $\mathcal{B}_i$, $\bm{V}_t$ is the updated cost-to-go value for global value iteration. If the algorithm is in phase $\mathcal{A}_i$, $\bm{V}_t$ represents $\tilde{\bm{V}}(\bm{W}_t)$.

\begin{theorem}\label{thm_adaptive}
 If $\limsup \alpha_t \to 0$, $\limsup_{t \to \infty} |\mathcal{A}_i| < \infty$  and  $\liminf_{t \to \infty} |\mathcal{B}_i| > 0$, we have 
$$
    \limsup_{t\rightarrow \infty}\|\bm{V}_t-\bm{V}^*\|_{\infty}
    \leq
    \frac{2\varepsilon}{1-\gamma}.
$$
\end{theorem}

Notice that the result of Theorem \ref{thm_adaptive} is consistent with Proposition \ref{prop_prefix}: due to state aggregation we suffer from the same $\frac{1}{1-\gamma}$ for the error bound. However, our algorithm maintains the same order of error bound without knowing the optimal value function $\bm{V}^*$. 

We also identify the existence of ``stable field'' for our algorithm, and we prove that under some specific choice of the learning rate $\alpha_t$, with probability one the value function will stay within the stable field.

\begin{proposition}\label{prop2}
    If $\alpha_t<\frac{\varepsilon}{(1-\gamma)\sup_{i}|\mathcal{A}_i|}$ for all $t$, we have that after $O\left(\max\left\{1,\frac{\log(\varepsilon)}{\log(\gamma)}\right\}\right)$ iterations, with probability one the estimated approximation $\bm{V}_t$ satisfies 
    $$
        \|\bm{V}_t-\bm{V}^*\|_{\infty}
        \leq
        \frac{3\varepsilon}{1-\gamma}
    $$
    for any choice of initialization $\bm{V}_0\in\left[0,\frac{1}{1-\gamma}\right]^{|\mathcal{S}|}$.
    % Here, the inequality holds entry-wisely.
    % HANS: This holds entrywise automatically, since the norm is the \(\ell^\infty\) norm, no?
\end{proposition}

\begin{proposition}\label{prop3}
For $\beta > 0$, if $\alpha_t \leq t^{-\beta}$ , after $O( \frac{\log(\varepsilon)}{\log(\gamma)} + (1-\gamma)^{-\frac{1}{\beta}}\varepsilon^{-\frac{1}{\beta}})$ iterations, with probability one the estimated approximation $\bm{V}_t$ satisfies 
    $$
        \|\bm{V}_t-\bm{V}^*\|_{\infty}
        \leq
        \frac{3\varepsilon}{1-\gamma}
    $$
for any choice of initialization $\bm{V}_0\in\left[0,\frac{1}{1-\gamma}\right]^{|\mathcal{S}|}$.
\end{proposition}

% With a little abuse of notation, we denote $\mathbb{E}_{s_0\sim \rho}[V^{\pi}(s_0)]$ the expected discounted sum of future rewards.
Such results provide guidance for choice of parameters. Indeed, the experimental results are consistent with the theory presented above.

\begin{figure}[t]
\centering
\begin{subfigure}{.32\textwidth}
  \centering
  \includegraphics[width=.6\linewidth]{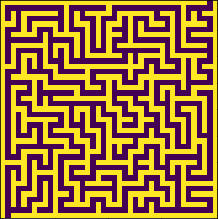}
  \caption{Standard Maze}
  \label{fig:ex-standard}
\end{subfigure}
\begin{subfigure}{.32\textwidth}
  \centering
  \includegraphics[width=.6\linewidth]{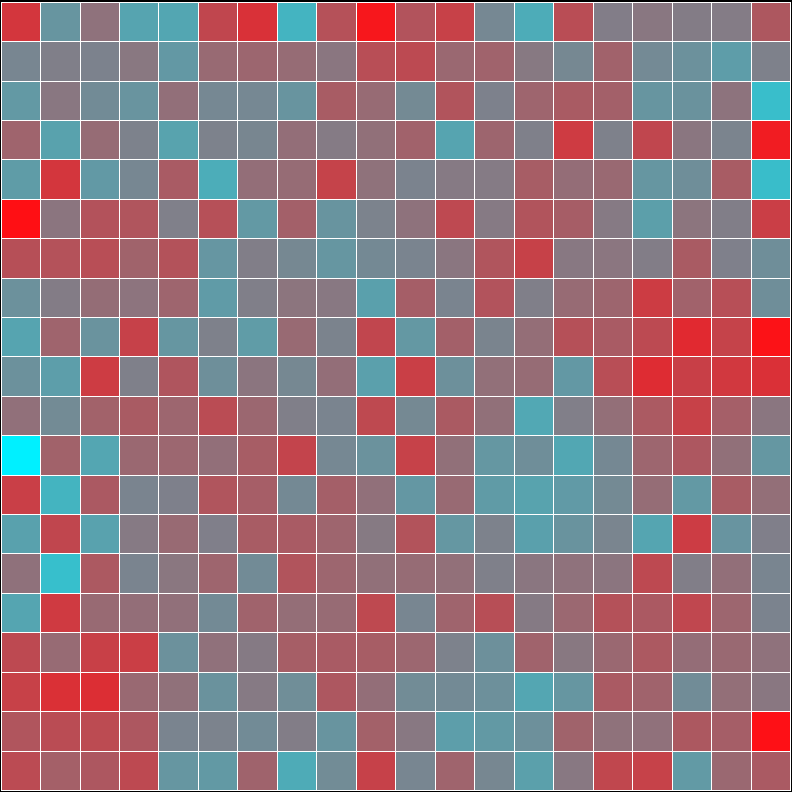}
  \caption{Terrain Maze}
  \label{fig:ex-terrain}
\end{subfigure}
\begin{subfigure}{.32\textwidth}
  \centering
  \includegraphics[width=.8\linewidth]{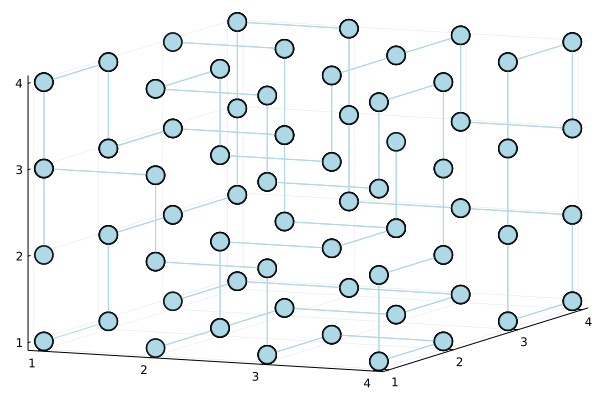}
  \caption{3-Dimensional Standard Maze}
  \label{fig:ex-3d}
\end{subfigure}%
\label{fig:ex}
\caption{
    \textbf{Left:} In the standard maze, the player's objective is to navigate to the bottom left. \textbf{Middle:} In the terrain maze, the player proceeds to the bottom left corner. Greater costs are
    incurred for moving uphill than downhill. High positions are indicated by red colors, and
    low positions are indicated by blue colors. \textbf{Right:} An example multi-dimensional problem.
    Here too the player's objective is to navigate to the lower-left corner (i.e., \((1,1,1)\)).
%We tested our method on high dimensional problems to showcase its ability to scale}
}
\end{figure}

\section{Experiments}
\label{sec:exp}

To test the theory developed in Sections \ref{sec:sa}, we perform a number of numerical experiments.\footnote{%
    All experiments were performed in parallel using forty Xeon E5-2698 v3 @ 2.30GHz CPUs. Total compute time was approximately 60 hours. Code and replication materials are available in the Supplementary Materials.
} We test our methods on a variety of MDPs of different sizes and complexity.
Our results show that state aggregation can achieve faster convergence than standard value iteration;
that state aggregation scales appropriately as the size and dimensionality of the underlying MDP increases;
and that state aggregation is reasonably robust to measurement error (simulated by adding noise to the action costs) and varying levels of stochasticity in the transition matrix.

%In this section we showcase our experimental results. We first run our method on 2-dimensional deterministic environments of varying size to first show that our method works. We then run our method on higher dimensional MDPs to showcase the scalability of our algorithm. We conclude our experiments with robustness tests including stochastic MDPs and added noise to cost vectors to further solidify our algorithm as a practical general method.

\begin{figure}[t]
\centering
\begin{subfigure}{0.32\textwidth}
  \centering
    \includegraphics[width=.9\linewidth]{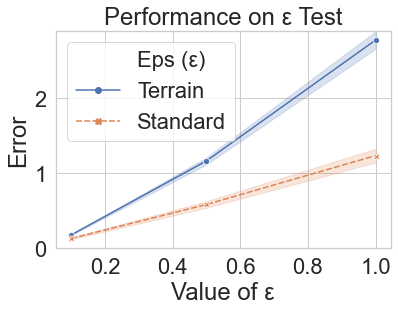}
    \caption{Influence of \(\varepsilon\)}
    \label{fig:eps_test}
\end{subfigure}%
\begin{subfigure}{0.33\textwidth}
  \centering
    \includegraphics[width=.9\linewidth]{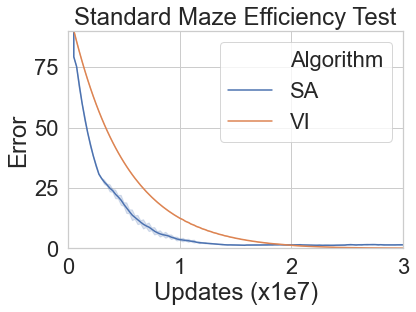}
    \caption{Efficiency test on standard maze}
    \label{fig:efficient_test_S}
\end{subfigure}
\begin{subfigure}{0.33\textwidth}
  \centering
    \includegraphics[width=.9\linewidth]{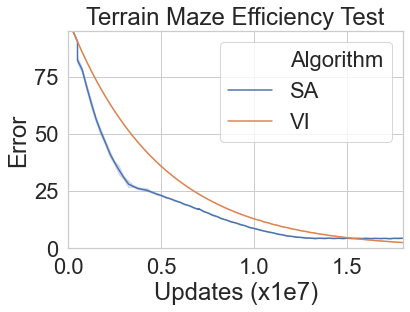}
    \caption{Efficiency test on terrain maze}
    \label{fig:efficient_test_T}
\end{subfigure}
\caption{
    \textbf{Left:} The error of state-aggregated value iteration after convergence as a function of \(\varepsilon\).
    \textbf{Middle:} Average convergence speed and 95\% confidence intervals of state-aggregated value iteration on \(500 \times 500\) standard mazes.
    \textbf{Right:} Average convergence speed and 95\% confidence intervals of state-aggregated value iteration on \(500 \times 500\) terrain mazes.
}
\label{fig:test}
\end{figure}

\subsection{MDP problems}

We consider two problems in testing our algorithm. The first, which we term the ``standard maze problem'' consists of a \(d_1 \times d_2 \times \cdots d_n\) grid of positions.
Each position is connected to one or more adjacent positions. Moving from position to position incurs a constant cost, except for moving to the terminal state of the maze (the position \((1, \ldots, 1)\)) which incurs a constant reward.\footnote{
    We rescale action costs in both the standard and terrain mazes to ensure that the maximum cost-to-go is exactly 100.
}
There is a unique path from each position to the terminal state.
A two-dimensional \(20 \times 20\) standard maze, in which the player can move, depending on their position, up, down left, or right is illustrated in Figure~\ref{fig:ex-standard}.

%The standard maze problem and the terrain maze problem. The base standard maze game consists of an n x m 2-dimensional maze  maze. Each state represents a position in the maze. We imagine that the player can move from state to state (that is, up, down, left, or right) as allowed by the constraints of the maze. We further require that each action incurs an immediate cost of 1, while the player can obtain a unit reward (i.e., a cost of -1) by reaching the final square of the maze, which
%by convention we place at position (n, m).

The second problem is the ``terrain maze problem.'' As in a standard maze, each state in the terrain maze represents a position in a \(d_1 \times \cdots \times d_n\) grid. As before, we imagine that the player can move from state to state only by travelling a single unit in any direction. (The player is only constrained from moving out of the grid entirely.) The player receives a reward
for reaching the final square
of the maze, which again we place at position \((1, \ldots, 1)\). However, in contrast to the standard
maze game, the player’s movements incur different costs at different positions. In particular, the maze is determined by a ``height function'' $H : \{1, \ldots, d_1\} \times \cdots \times \{1, \ldots, d_n\} \to \mathbb{R}$. The cost of movement is set to be the difference in heights between the player’s destination position and their current position, normalized appropriately.

In both problems, we also allow for stochasticity controlled by a parameter \(p\) which gives the probability that a player moves in their intended direction. For \(p = 1\), the MDP is deterministic; otherwise, with probability \(1 - p\), the player moves in a different direction chosen uniformly at random.

%We use these two problems because they have very different properties and help illustrate that our method is able to generalize to a wide variety of different MDP problems.
In both problems, the actions available at any state correspond to a very sparse transition probability vectors, since players are constrained to move along cardinal directions at a rate of a single unit. However, in the standard maze game, the cost-to-go at any position is extremely sensitive to the costs-to-go at states that are very distant. In a \(10 \times 10\) standard maze, the initial tile (i.e., position (1, 1)) is often between 25 and 30 units away from the destination tile (i.e., position (10, 10)). %\hans{I kind of want to delete the second-half of this paragraph. It doesn't feel rigorous enough for this setting.} However, the cost-to-go is essentially a function of that distance, and so accurately updating the cost-to-go of the initial tile depends on accurately updating the costs-to-go at all of the 25 to 30 tiles between it and the destination tile.
In contrast, in the terrain maze game, the cost-to-go is much less sensitive to far away positions, because local immediate costs, dictated by the slopes one must climb or go down to move locally, are more significant.

%\subsection{Significant Speedups}
%A significant contribution for having a feasible state aggregation algorithm is the speedup in convergence on solving MDP problems. Before presenting our results, we define a key metric.

%\textbf{State-Aggregation Convergence:} The state-aggregation convergence is the convergence reached by the state aggregation method when given the true v-values to start with.

%The reason we use this metric in our experiments is given the true v-values, our method cannot possibly beat the convergent values achieved when run on the true v-values as initial guesses. The convergent values achieved are very close to the true values and are approximate enough to be feasible for most real world problems where the problem is so large that regular value iteration would be infeasible to use. The advantage here is that state aggregation offers such a speedup that the approximate convergent values are good enough. We argue then that upon reaching the state-aggregation convergence values, our method has finished.

%We present numerical results for our method on the two MDP problems in the standard and terrain mazes (in reaching State-Aggregation Convergence2) to showcase the impressive speedup our algorithm offers.

\subsubsection{Benchmarks and parameters}

We measure convergence by the \(\ell^\infty\)-distance (hereafter ``error'') between the current cost-to-go vector and the true cost-to-go vector, and we evaluate the speed based on the size of error and the number of updates performed. Notice that for global value iteration, an update for state $s$ has the form 
$$T_s(\bm{V}) = \min_{a\in \mathcal{A}}\left(r(s, a) + \gamma \bm{P}_{s,a}^{\top} \bm{V}\right),$$
and an update for mega-state $j$ (represented by $s$) has the form
$$W_t(j) = (1-\alpha_{t-t_0})W_{t-1}(j) + \alpha_{t-t_0} T_s\tilde{\bm{V}}(\bm{W}_{t-1}).$$
Because the transition matrix is not dense in our examples, the computational resources required for global value iteration update and aggregated update are roughly the same. For each iteration, the value iteration will always perform $|\mathcal{S}|$ updates, and for Algorithm \ref{alg:AVIA}, only
%$\|\bm{V}\|_{\infty}/\varepsilon$
\(K\) updates, one for each mega-state, will be performed if in the aggregation phase.

All experiments are performed with a discount factor \(\gamma = 0.95\). We set $|\mathcal{A}_i| = 5$ and $|\mathcal{B}_i| = 2$ for every $i$, and for learning rate we set $\alpha_t = \frac{1}{\sqrt{t}}$ . The cost function is normalized such that $\|\bm{V}^*\|_{\infty} = 100$, and we choose aggregation constant to be \(\varepsilon = 0.5\) (unless otherwise indicated). We set the initial cost vector \(\bm{V}_0\) to be the zero vector \(\bm{0}\).

\subsubsection{Results}

\paragraph{Influence of \(\varepsilon\).}

We test the effect of $\varepsilon$ on the error Algorithm~\ref{alg:AVIA} produces. We run experiments with aggregation constant $\varepsilon = 0.05, 0.1$, and $ 0.5$ for $500 \times 500$ standard and terrain mazes. For each $\varepsilon$, we run 1,000 iterations of Algorithm \ref{alg:AVIA}, and each experiment is repeated 20 times; the results, shown in Figure~\ref{fig:eps_test}, indicate that the approximation error scales in proportion to \(\varepsilon\), which is consistent with Proposition \ref{prop_prefix} and Theorem \ref{thm_adaptive}.

\paragraph{Efficiency.}

We test the convergence rate of Algorithm \ref{alg:AVIA} against value iteration on \(500 \times 500\) standard and terrain mazes, repeating each experiment 20 times. From Figure~\ref{fig:efficient_test_S}~and~\ref{fig:efficient_test_T}, we see that state-aggregated value iteration is very efficient at the beginning phase, converging in fewer updates than value iteration.

%we run 20 numerical experiment to contruct the confidence interval.
%\textbf{\textcolor{blue}{Let $\varepsilon = 0.05$, $\varepsilon = 0.1$, $\varepsilon = 0.5$. Need a graph with confidence interval (see Abel 20)}} 

% \begin{table}
%   \caption{Sample table title}
%   \label{sample-table}
%   \centering
%   \begin{tabular}{lll}
%     \toprule
%     \multicolumn{2}{c}{Part}                   \\
%     \cmidrule(r){1-2}
%     Name     & Description     & Size ($\mu$m) \\
%     \midrule
%     Dendrite & Input terminal  & $\sim$100     \\
%     Axon     & Output terminal & $\sim$10      \\
%     Soma     & Cell body       & up to $10^6$  \\
%     \bottomrule
%   \end{tabular}
% \end{table}

% \begin{table}[!h]
%   \caption{Performance on $\varepsilon$ test}
%   \label{sample-table}
%   \centering
%   \begin{tabular}{lll}
%     \toprule
%     $\varepsilon$     & Mean Err    & 95 CI \\
%     \midrule
%     T $0.1$ & 0.179  & (0.145, 0.213) \\
%     T $0.5$   & 1.166 & (0.91, 1.42)  \\
%     T $1$   & 2.778  & (2.145, 3.411)  \\
%     S $0.1$ & 0.131  & (0.064, 0.199) \\
%     S $0.5$   & 0.585 & (0.345, 0.824)  \\
%     S $1$   & 1.236  & (0.735, 1.738)  \\
%     \bottomrule
%   \end{tabular}
% \end{table}

\begin{table}
  \parbox{.45\linewidth}{
  \centering
  \begin{tabular}{lrrr}
    \toprule
    Type & Dims.     & Error     & 95\% CI \\
    \midrule
    Trn. & $100 \times 100$ & 4.41 & $\pm 0.14$ \\
    Trn. & $200 \times 200$ & 4.34 & $\pm 0.16$ \\
    Trn. & $300 \times 300$ & 4.65 & $\pm 0.18$ \\
    Trn. & $500 \times 500$ & 4.27 & $\pm 0.17$ \\
    Trn. & $1000 \times 1000$ & 4.27 & $\pm 0.16$ \\
    Std. & $100 \times 100$ & 1.43 & $\pm 0.16$ \\
    Std. & $200 \times 200$ & 1.39 & $\pm 0.15$ \\
    Std. & $300 \times 300$ & 1.42 & $\pm 0.20$ \\
    Std. & $500 \times 500$ & 1.11 & $\pm 0.16$ \\
    Std. & $1000 \times 1000$ & 1.40 & $\pm 0.16$ \\
    \bottomrule
  \end{tabular}
  }
  \hfill
  \parbox{.45\linewidth}{
  \centering
  \begin{tabular}{lrrr}
    \toprule
    Type & Dims.     & Error     & 95\% CI \\
    \midrule
    Trn. & $10^3$    & 1.91  & $\pm 0.007$ \\
    Trn. & $10^4$    & 3.02  & $\pm 0.009$ \\
    Trn. & $10^5$    & 3.59  & $\pm 0.008$ \\
    Trn. & $10^6$    & 3.85  & $\pm 0.005$ \\
    Std. & $10^3$    & 1.36  & $\pm 0.019$ \\
    Std. & $10^4$    & 1.36  & $\pm 0.017$ \\
    Std. & $10^5$    & 1.23  & $\pm 0.013$ \\
    Std. & $10^6$    & 1.31  & $\pm 0.013$ \\
    \bottomrule
  \end{tabular}
  }
  \medskip
    \caption{
        Scaling properties of state aggregation value iteration. Reported errors represent the mean value from running each experiment 20 times.
    }
    \label{tbl:nd}
\end{table}

\begin{table}
  \parbox{.49\linewidth}{
  \centering
  \begin{tabular}{lrrrr}
    \toprule
    Type & $p$ & $\sigma$ & Error & 95\% CI  \\
    \midrule
    Terrain & 0.92 & 0.00 & 4.44 & $\pm 0.24$ \\
    Terrain & 0.92 & 0.01 & 4.41 & $\pm 0.18$ \\
    Terrain & 0.92 & 0.05 & 4.97 & $\pm 0.17$ \\
    Terrain & 0.92 & 0.10 & 6.36 & $\pm 0.19$ \\
    Terrain & 0.95 & 0.00 & 4.43 & $\pm 0.17$ \\
    Terrain & 0.95 & 0.01 & 4.32 & $\pm 0.14$ \\
    Terrain & 0.95 & 0.05 & 4.93 & $\pm 0.17$ \\
    Terrain & 0.95 & 0.10 & 6.39 & $\pm 0.17$ \\
    Terrain & 0.98 & 0.00 & 4.37 & $\pm 0.19$ \\
    Terrain & 0.98 & 0.01 & 4.31 & $\pm 0.14$ \\
    Terrain & 0.98 & 0.05 & 5.01 & $\pm 0.14$ \\
    Terrain & 0.98 & 0.10 & 6.52 & $\pm 0.20$ \\
    \bottomrule
  \end{tabular}
  }
  \hfill
  \parbox{.49\linewidth}{
  \centering
  \begin{tabular}{lrrrr}
    \toprule
    Type & $p$ & $\sigma$ & Error & 95\% CI  \\
    \midrule
    Standard & 0.92 & 0.00 & 1.39 & $\pm 0.19$ \\
    Standard & 0.92 & 0.01 & 1.61 & $\pm 0.23$ \\
    Standard & 0.92 & 0.05 & 2.62 & $\pm 0.15$ \\
    Standard & 0.92 & 0.10 & 5.56 & $\pm 0.46$ \\
    Standard & 0.95 & 0.00 & 1.49 & $\pm 0.17$ \\
    Standard & 0.95 & 0.01 & 1.57 & $\pm 0.14$ \\
    Standard & 0.95 & 0.05 & 2.86 & $\pm 0.17$ \\
    Standard & 0.95 & 0.10 & 5.72 & $\pm 0.39$ \\
    Standard & 0.98 & 0.00 & 1.43 & $\pm 0.18$ \\
    Standard & 0.98 & 0.01 & 1.88 & $\pm 0.16$ \\
    Standard & 0.98 & 0.05 & 3.48 & $\pm 0.28$ \\
    Standard & 0.98 & 0.10 & 5.86 & $\pm 0.26$ \\
    \bottomrule
  \end{tabular}
  }
  \medskip
    \caption{
         Numerical experiments illustrating the robustness of state-aggregated value iteration to stochasticity and noisy action costs. Errors represent the average \(\ell_\infty\)-distance to the true cost-to-go values in twenty independent runs.
    }
    \label{tbl:robust}
\end{table}

%\textbf{\textcolor{red}{Here it would be nice to have results on 2-d examples of varying sizes perhaps 50x50, 250x250, 500x500 the sizes chosen illustrate both small and large examples for both standard maze and terrain}} 

%\textbf{\textcolor{blue}{Fix $\varepsilon = 0.1$, maybe use a table instead of chart make more sense?}} 

\paragraph{Scalibility of state aggregation.}
%We now run on very high dimensional MDP problems with large state and action spaces to show the significance of our algorithm.
We run state-abstracted value iteration on standard and terrain mazes of size \(100 \times 100\), \(200 \times 200\), \(300 \times 300\), \(500 \times 500\), and \(1000 \times 1000\) for 1000 iterations. We repeat each experiment 20 times, displaying the results in Table~\ref{tbl:nd}. Next, we run state-aggregated value iteration on terrain mazes of increasingly large underlying dimension, as shown in the right side of Table~\ref{tbl:nd}, likewise for 20 repetitions for each size. The difference with the previous experiment is that not only does the state space increase, the action space also increases exponentially. Our results show that the added complexity of the high-dimensional problems does not appear to substantially affect the convergence of state-aggregated value iteration and our method is able to scale with very large MDP problems.

\paragraph{Robustness.}

We examine the robustness of state-aggregated value iteration to two sources of noise. We generate \(500 \times 500\) standard and terrain mazes, varying the level of stochasiticity by setting \(p = 0.92, 0.95, 0.98\). We also vary the amount of noise in the cost vector by adding a mean 0, standard deviation \(\sigma = 0.0, 0.01, 0.05, 0.1\) normal vector to the action costs. The results, shown in Table~\ref{tbl:robust}, indicate that state-aggregated value iteration is reasonably robust to stochasiticity and measurement error.

\subsection{Continuous Control Problems}
We conclude the experiments section by showing the performance of our method on a real-world use case in continuous control. These problems often involve solving complex tasks with high-dimensional sensory input. The idea typically involves teaching an autonomous agent, usually a robot, to successfully complete some task or achieve some goal state. These problems are often very tough as they reside in the continuous state space (and many times action space) domain. 

We already showcased the significant reduction in algorithm update costs during learning on grid-world problems in comparison with value iteration. Our goal for this section is to showcase real world examples of how our method may be practically applied in the field of continuous control. This is important not only to emphasize that our idea has a practical use case, but also to further showcase its ability to scale into extremely large (and continuous) dimensional problems. 

\subsubsection{Environment}

We choose to use the common baseline control problem in the "CartPole" system. The CartPole system is a typical physics control problem where a pole is attached to a central joint of a cart, which moves along an endless friction-less track. The system is controlled by applying a force to either move to the right or left with the goal to balance the pole, which starts upright. The system terminates if (1) the pole angle is more than $12$ degrees from the vertical axis, or (2) the cart position is more than $2.4$ cm
from the center, or (3) the episode length is greater than
$200$ time-steps. 

During each episode, the agent will receive constant reward for each step it takes. The state-space is the continuous position and angle of the pole for the cartpole pendulum system and the action-space involves two actions - applying a force to move the cart left or right. A more in-depth explanation of the problem can be found in \cite{DBLP:journals/corr/abs-2006-04938}.

\subsubsection{Results}

Since the CartPole problem is a multi-dimensional continuous control problem, there is no ground truth $v$-values so we choose to utilize the dense reward nature of this problem and rely on the accumulated reward to quantify algorithm performance as opposed to error. In addition, since value iteration requires discrete states, we discretize all continuous dimensions of the state space into bins and generate policies on the discretized environment. More specifically, we discretize the continuous state space into bins by dividing each dimension of the domain into equidistant intervals.

\begin{figure}[t]
\centering
\begin{subfigure}{0.47\textwidth}
  \centering
    \includegraphics[width=.9\linewidth]{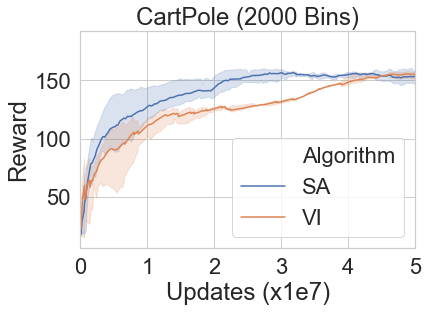}
    % \caption{Discretization: }
    \label{cartexp}
\end{subfigure}%
\begin{subfigure}{0.47\textwidth}
  \centering
    \includegraphics[width=.9\linewidth]{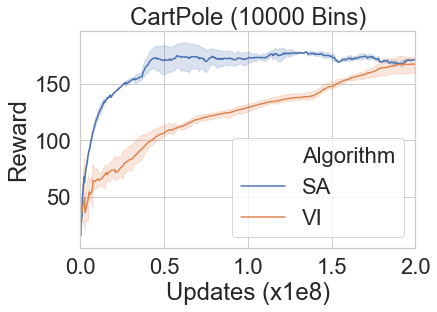}
    % \caption{Discretization: }
    \label{cartexp2}
\end{subfigure}
\caption{
    CartPole problem reward versus number of updates comparison between value iteration and our adaptive state aggregation method. Even in use-cases such as continuous control, our method offers valuable speedups in the learning process of MDPs and stays robust against different discretization schemes.
}\label{fig:cartexp}
\end{figure}

For our aggregation algorithm, we use  $\gamma = 0.99$ following what is commonly used in this problem in past works. We set the initial cost vector $V_0$ to be the zero vector. We fix $\varepsilon=0.2$ and set $\alpha_t = \frac{1}{\sqrt{t}}$. We note that given the symmetric nature of this continuous control problem, we do not need to alternate between global and aggregate updates. The adaptive aggregation of states already groups states effectively together for strong performance and significant speedups in the number of updates required. We first do a sweep of number of bins to discretize the problem space to determine the number that best maximizes the performance of value iteration in terms of both update number and reward and found this to be around 2000 bins. We then compare the performances of state aggregation and value iteration on this setting in figure \ref{fig:cartexp}. To also further showcase the adaptive advantage of our method, we choose a larger bin (10000) number that likely would have been chosen if no bin sweeping had occurred and show that our method acts as an "automatic bin adjuster" and offers the significant speedups without any prior tuning in figure \ref{fig:cartexp}. Interestingly in both situations, the experimental results show that the speedup offered by our method seem significant across different bin settings of the CartPole problem. These results may prove to have strong theoretical directions in future works.

\section{Discussion}

Value iteration is an effective tool for solving Markov decision processes but can quickly become infeasible
%for use
in problems with large state and action spaces.
To address this difficulty, we develop adaptive state-aggregated value iteration, a novel method of solving Markov decision processes by aggregating states with similar cost-to-go values and updating them in tandem.
Unlike previous methods of reducing the dimensions of state and action spaces, our method is general and does not require prior knowledge of the true cost-to-go values to form aggregate states.
Instead our algorithm learns the cost-to-go values online and uses them to form
%these
aggregate states in an adaptive manner.
We prove theoretical guarantees for the accuracy, convergence, and convergence rate of our state-aggregated value iteration algorithm,
and demonstrate its applicability through a variety of numerical experiments.

State- and action-space reduction techniques are an area of active research. Our contribution in the dynamic state-aggregated value iteration algorithm provides a general framework for approaching large MDPs with strong numerical performances justifying our method. We believe our algorithm can serve as a foundational ground for
%future works including interesting theoretical directions along with new empirical methods.
both future empirical and theoretical work.

We conclude
%our paper
by discussing
%areas of
promising directions
%to further improve upon our adaptive state-aggregation idea.
for future work on adaptive state aggregation.
% In terms of the limitation of the work and possible improvement. 
First, we believe that reducing the number of updates per state by dynamically aggregating states can also be extended more generally in RL settings with model-free methods.
Second, our proposed algorithm's complexity is of the same order as value iteration.
Future work may seek to eliminate the dependence on $\gamma$ in the error bound.
Lastly, by adaptively choosing $\varepsilon$,
it may be possible to achieve better complexity bounds not only for planning problems but also for generative MDP models \cite{sidford2018,sidford2018near}.

%\textbf{\textcolor{red}{Here it would be nice to have results on 2-d examples of fixed dimension size perhaps 100x100 or 200x200 (should be large enough to be challenging) and run the stochastic and noisy settings}}

%\textbf{\textcolor{red}{All runs should be run 3 times and averaged and plotted with error bars}}

%\textbf{\textcolor{blue}{actually, can we run it 20 times with and without random noise on reward and varying p to construct the confidence interval. Fix epsilon to be $0.1$ and a run 100 x 100 example. }} 

\bibliographystyle{abbrv} % outcomment this and next line in Case 1
\bibliography{main}

\appendix

\section{Appendix}
\subsection{Preliminaries}
Since the number state-action pairs is finite, without loss of generality we can assume that the cost function $r$ is bounded by $1$. By a slight abuse of notation, we denote by $\bm{V}_t$ the current cost-to-go value at iteration $t$ generated by Algorithm \ref{alg:AVIA}.
More specifically, if the current algorithm is in global value iteration phase $\mathcal{B}_i$, $\bm{V}_t$ is the updated cost-to-go value for global value iteration. If the algorithm is in state aggregation phase $\mathcal{A}_i$, $\bm{V}_t$ represents $\tilde{\bm{V}}(\bm{W}_t)$. Also, denote by $\bm{E}_t = \bm{V}_t - \bm{V}^*$ the error between the current estimate and the optimal value function. We first state results on the \(\ell^\infty\) norm of $\bm{V}^*$ and $\bm{V}_t$.

\begin{lemma}\label{lem_bdd}
For any $t > 0$ we have $||\bm{V}_t||_{\infty} \leq \frac{1}{1-\gamma}$ and $||\bm{V}^*||_{\infty} \leq \frac{1}{1-\gamma}$.
\end{lemma}

Lemma \ref{lem_bdd} states that both of the optimal cost-to-go value and the value obtained by Algorithm \ref{alg:AVIA} is uniformly bounded by $\frac{1}{1-\gamma}$ entry-wisely at all aggregated and global iterations. 

Next, it is well known that $||\bm{E}_t||_{\infty}$ can only decrease in the global value iteration phase due to the fact that the Bellman operator is a contraction mapping. However, $||\bm{E}_t||_{\infty}$ can potentially grown when we switch into a state aggregation phase. To evaluate the fluctuation of $||\bm{E}_t||_{\infty}$ during the state aggregation phase, we firstly notice that the step 4 in Algorithm \ref{alg:A} (which is the initialization step for state aggregation) will generate a tight bound: if we denote $\bm{W}(\bm{V}_t)$ to be the initialization result in step 4 Algorithm \ref{alg:A}, we will have 
\begin{align}\label{ineq_error_1}
||\tilde{\bm{V}}(\bm{W}(\bm{V}_t)) - \bm{V}^*||_{\infty} \leq ||\bm{V}_t - \bm{V}^*||_{\infty} + \frac{\varepsilon}{2}.
\end{align}
The inequality above is tight because by construction there exists a scenario such that $\bm{V}_t = \bm{V}^*$ and $\tilde{\bm{V}}(\bm{W}(\bm{V}_t))(s)-\bm{V}^*(s) = \frac{\varepsilon}{2}$ for some $s \in \mathcal{S}$.

After the error introduced in the initialization stage of stage aggregation, the value updates process may also accumulate error. Recall that during each state aggregation phase $\mathcal{A}_i$, for $t \in \mathcal{A}_i$ we have the following updates for mega-state $j$
$$ W_t(j) = (1-\alpha_{t_{sa}})W_{t-1}(j) + \alpha_{t
        _{sa}} T_s\tilde{\bm{V}}(\bm{W}_{t-1}).$$
Hence by combining Lemma \ref{lem_bdd}, we know that for every SA iteration we have
$$|W_t(j) - W_{t-1}(j)| \leq  \alpha_{t
        _{sa}}( ||T_s\tilde{\bm{V}}(\bm{W}_{t-1})||_{\infty} + ||\bm{W}_{t-1})||_{\infty}) \leq \alpha_{t
        _{sa}}\frac{2}{1-\gamma}.$$

We then state a general condition that will control the error bound of the state aggregation value update, and we will show that all of our theorem and propositions satisfy such condition.
\begin{condition}\label{cdn}
There exists $M > 0$ such that for any state aggregation phase $\mathcal{A}_i$ coming after $M$ state aggregation iterations, we have
\begin{equation}\label{ineq_error_2_general}
    \begin{aligned}
    ||\bm{W}_{t_0 + |\mathcal{A}_i|} - \bm{W}_{t_0}||_{\infty} \leq  \sum_{i=0}^{|\mathcal{A}_i|-1}\alpha_{t
        _{sa}}( ||T_s\tilde{\bm{V}}(\bm{W}_{t_0+i})||_{\infty} + ||\bm{W}_{t_0+i}||_{\infty}) \leq \frac{\varepsilon}{2}.
    \end{aligned}
\end{equation}
\end{condition}
During state aggregation phase, from $\eqref{ineq_error_1}$ we know that the if we switch from $\mathcal{B}_i$ to $\mathcal{A}_i$, $\bm{E}_t$ can increase by $\frac{\varepsilon}{2}$, and from Condition \ref{cdn} we know that after $|\mathcal{A}_i|$ steps of value iteration, $\bm{E}_t$ can increase by another $\frac{\varepsilon}{2}$.

More specifically, after $M$ iterations suggested by Condition \ref{cdn}, if we take $t_0 = \max\{t \mid t \in \mathcal{B}_i\}$ and $t_1 = \min\{t \mid t \in \mathcal{B}_{i+1}\}$, from our discussion above we have
\begin{equation}\label{ineq_error_general}
    \begin{aligned}
    ||\bm{V}_{t_1} - \bm{V}^*||_{\infty} \leq ||\bm{V}_{t_0} - \bm{V}^*||_{\infty} + \varepsilon.
    \end{aligned}
\end{equation}
Moreover, the above equation also holds for all $t \in \mathcal{B}_{i+1}$, as the Bellman operator is a contraction.

\begin{figure}[t]
\centering
    \includegraphics[width=1\linewidth]{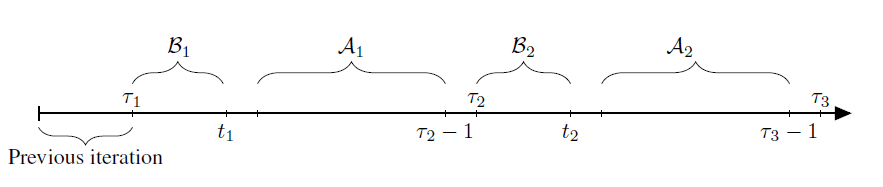}
    \caption{Notation after re-indexing}
\label{fig:proof}
\end{figure}

Then, for notational convenience, we re-index the first global iteration phase after $M$ SA iterations to be $\mathcal{B}_1$, the following time interval to be $\mathcal{A}_1, \mathcal{B}_2, \mathcal{A}_2, \cdots$. We also denote $\tau_i = \min\{t \mid t\in \mathcal{B}_i\}$ and $t_i = \max\{t \mid t\in \mathcal{B}_i\}$; see Figure \ref{fig:proof}.

Since $\lim\inf |\mathcal{B}_i| \geq 1$, there exists $I > 0$ such that for any $i > I$ and $t \in (t_i, \tau_{i+1} - 1]$ we have
\begin{equation}\label{ineq_recur}
    \begin{aligned}
    ||\bm{V}_{t_i} - \bm{V}^*||_{\infty} &\leq \gamma||\bm{V}_{\tau_i - 1} - \bm{V}^*||_{\infty}\\
    ||\bm{V}_{t} - \bm{V}^*||_{\infty} &\leq ||\bm{V}_{t_i} - \bm{V}^*||_{\infty} + \varepsilon,
    \end{aligned}
\end{equation}
where the first inequality comes from the fact that during $[\tau_i, t_i]$ the global value iteration is performed. The second inequality comes from combining $\eqref{ineq_error_1}$ and $\eqref{ineq_error_2_general}$.

The intuition of $\eqref{ineq_recur}$ is that from $\tau_{i} - 1$ to $t_i$, our approximation will only get better because we are using global value iteration only. From time $t_i$ to $\tau_{i+1} - 1$, the approximation within the entire SA value iteration phase will stay within the previous error up to another $\varepsilon$. By balancing those bounds we are able to get a ``stable field'' for all phases. 

We then state the bounds for the error at $t_{i}$ for $i > 0$
\begin{lemma}\label{lem}
For all $n \geq 1$, we have 
\begin{align}
    ||\bm{V}_{t_{n+1}} - \bm{V}^*||_{\infty} \leq \gamma^{n}||\bm{V}_{\tau_1} - \bm{V}^*||_{\infty} + \frac{\varepsilon}{1-\gamma}.
\end{align}
\end{lemma}
From Lemma \ref{lem} and $\eqref{ineq_recur}$, we can generate bounds for all $t$ that is large.

\subsection{Proof of Theorem 1}
Since $$\limsup_{t \to \infty} |\mathcal{A}_t| < \infty,\ \liminf_{t \to \infty} |\mathcal{B}_t| > 0,$$ 
there exists $N$ such that we have that when $i > N$,  $|\mathcal{A}_i|$ will be bounded from above and $|\mathcal{B}_i| $ will be bounded from below. We denotes those two constants for the bound as $a$ and $b$, respectively. Then, since the sequence $\{\alpha_t\}_{t=1}^{\infty}$ goes to zero as $t$ goes to infinity, there exists a constant $T$ such that for $t>T$ we have 
$a \cdot \alpha_t < \frac{\varepsilon}{2(1-\gamma)}$. Thus, we have
$$    ||\bm{W}_{t_0 + |\mathcal{A}_i|} - \bm{W}_{t_0}||_{\infty} \leq  \sum_{i=0}^{a}\alpha_{t
        _{sa}}( ||T_s\tilde{\bm{V}}(\bm{W}_{t_0+i})||_{\infty} + ||\bm{W}_{t_0+i}||_{\infty}) \leq \frac{\varepsilon}{2}.$$
Therefore, we have shown that Condition \ref{cdn} has been satisfied.

Next, we show that $\bm{V}_{t_i}$ grows nearer to $\bm{V}^*$ as $t$ increases. To see this, by directly applying Lemma \ref{lem}, we have
$$
\|\bm{V}_{t_i}-\bm{V}^*\|_{\infty}
\leq
\gamma^{i-1}\|\bm{V}_{t_1}-\bm{V}^*\|_{\infty} + 
\frac{\varepsilon}{1-\gamma},
$$
and, thus,
$$\lim\sup_{i\to\infty}\|\bm{V}_{t_i}-\bm{V}^*\|_{\infty}\leq\frac{\varepsilon}{1-\gamma}.$$
Then, in order to prove the final statement of the theorem, we aim to show that for any positive constant $\lambda>0$, there exists a constant $T_{\lambda}$ so that for any $t>T_{\lambda}$, the following inequality holds
\begin{align}\label{ineq_thm1_toshow}
\|\bm{V}_t-\bm{V}^*\|_{\infty}
\leq
\frac{2\varepsilon}{1-\gamma}
+
\lambda.
\end{align}

Since $\lim\sup\limits_{i\rightarrow\infty}\|\bm{V}_{t_i}-\bm{V}^*\|_{\infty}\leq\frac{\varepsilon}{1-\gamma}$, we can find $i_{\lambda} > 0$ such that for any $i\geq i_\lambda$, we have 
\begin{align}
\label{eqn:lims}
\|\bm{V}_{t_i}-\bm{V}^*\|_{\infty}\leq\frac{\varepsilon}{1-\gamma}+\lambda.
\end{align}
Then, by defining $T_\lambda=T+t_{i_\lambda}$, it suffices to  show Eq.~\eqref{ineq_thm1_toshow} for $t \in [T_\lambda, +\infty) \cap \mathcal{A}_i$ and $t \in [T_\lambda, +\infty) \cap \mathcal{B}_i$ separately. 
If $t \in [T_\lambda, +\infty) \cap \mathcal{A}_i$ for some $i\in\mathbb{N}$, from inequality $\eqref{ineq_recur}$  we have
$$
\|\bm{V}_t-\bm{V}_{t_{i}}\|\leq\varepsilon.
$$
Thus, 
\begin{align*}
\|\bm{V}_t-\bm{V}^*\|_{\infty}
&\leq
\|\bm{V}_t-\bm{V}_{t_{i}}\|_{\infty}
+
\|\bm{V}_{t_{i}}-\bm{V}^{*}\|_{\infty}\\
&\leq
\varepsilon+\frac{\varepsilon}{1-\gamma}
+
\lambda\\
&\leq
\frac{2\varepsilon}{1-\gamma}
        +
\lambda,
\end{align*}
where the first line is obtained by the triangle inequality and Eq.~\eqref{eqn:lims}. If $t \in [T_\lambda, +\infty) \cap \mathcal{B}_i$ for some $i\in\mathbb{N}$, based on the contraction property of Value-Iteration method, we have $\|\bm{V}_{t}-\bm{V}^*\|_{\infty}\leq \gamma^{t-\tau_i}\|\bm{V}_{\tau_i}-\bm{V}^*\|_{\infty}$. Hence, we have shown that Eq.~\eqref{ineq_thm1_toshow} holds for all $t > T_{\lambda}$ and any $\lambda > 0$, which is equivalent to
$$
        \lim\sup_{t\rightarrow\infty}\|\bm{V}_t-\bm{V}^*\|_{\infty}
        \leq
        \frac{2\varepsilon}{1-\gamma}.
$$

Moreover, if we set $\lambda=\varepsilon$ and the step size satisfies $\alpha_t<\frac{\varepsilon}{(1-\gamma)\sup_{i}|\mathcal{A}_i|}$ for all $t$, we have $T=1$ and $t_{i_{\lambda}}=O\left(\max\left\{1,\frac{\log(\varepsilon)}{\log(\gamma)}\right\}\right)$ . Then, we also can obtain Proposition \ref{prop2} using a similar argument.

\subsection{Proof of Proposition 3}

Without loss of generality, assume $|\mathcal{A}_i| = a $ and $ |\mathcal{B}_i| = b$ for all $i$. By taking $\alpha_t = t^{-\beta}$, after $\frac{(4a)^{1/\beta}}{(1-\gamma)^{1/\beta}\varepsilon^{1/\beta}}$ state aggregated iterations (which correspond to $\frac{(a+b)(4a)^{1/\beta}}{a(1-\gamma)^{1/\beta}\varepsilon^{1/\beta}}$ iterations in Algorithm \ref{alg:AVIA}), during any state aggregation phase $\mathcal{A}_i$ we have that $\sum_{t\in{\mathcal{A}_i}} \alpha_{t_{sa}} \leq \frac{(1-\gamma)\varepsilon}{4}$. 

Therefore, after the state aggregation phase $\mathcal{A}_i$, if we let $t_0 = \min_{t\in\mathcal{A}_i}$, we have
\begin{equation}\label{ineq_error_2}
    \begin{aligned}
    ||\bm{W}_{t_0 + a} - \bm{W}_{t_0}||_{\infty} \leq  \sum_{i=1}^a\alpha_{t
        _{sa}}( ||T_s\tilde{\bm{V}}(\bm{W}_{t_0})||_{\infty} + ||\bm{W}_{t_0-1}||_{\infty}) \leq \frac{\varepsilon}{2}.
    \end{aligned}
\end{equation}

Hence Condition \ref{cdn} is also satisfied. From Lemma \ref{lem}, we know that if we run more than $\frac{\log(\varepsilon)}{\log(\gamma)}$ global iteration phases (corresponds to $\frac{(a+b)\log(\varepsilon)}{\log(\gamma)}$ iterations in Algorithm \ref{alg:AVIA}), for the following $t_n$ we have

\begin{equation}\label{ineq_skeleton}
    \begin{aligned}
    ||\bm{V}_{t_{n}} - \bm{V}^*||_{\infty} \leq \gamma^{n-1}\frac{1}{1-\gamma} + \frac{\varepsilon}{1-\gamma} \leq \frac{2\varepsilon}{1-\gamma}.
    \end{aligned}
\end{equation}

Then, it suffices to prove that for all $t \geq t_n$, we have $ ||\bm{V}_{t_{n}} - \bm{V}^*||_{\infty} \leq \frac{3\varepsilon}{1-\gamma}$. We first check that the bound hold for state aggregated iterations. For any $i \geq n$ and any $t \in [t_i, \tau_{i+1} - 1]$, from $\eqref{ineq_recur}$ and $\eqref{ineq_skeleton}$ we know that
\begin{align*}
    ||\bm{V}_{t} - \bm{V}^*||_{\infty} &\leq ||\bm{V}_{t_i} - \bm{V}^*||_{\infty} + \varepsilon \leq \frac{3\varepsilon}{1-\gamma}.
\end{align*}
Lastly, for any $i > n$ and any $t \in [\tau_{i}, t_i]$, from contraction property we know that
\begin{align*}
    ||\bm{V}_{t} - \bm{V}^*||_{\infty} &\leq \gamma^{t-\tau_i+1}||\bm{V}_{\tau_{i} - 1} - \bm{V}^*||_{\infty} \leq \gamma^{t-\tau_i+1}\frac{3\varepsilon}{1-\gamma} \leq \frac{3\varepsilon}{1-\gamma}.
\end{align*}
Therefore, the $\bm{V}_t$ process will stay stable for $t \geq t_n$, where $t_n = \frac{(a+b)(4a)^{1/\beta}}{a(1-\gamma)^{1/\beta}\varepsilon^{1/\beta}} + \frac{(a+b)\log(\varepsilon)}{\log(\gamma)}$.

\subsection{Proof of Lemma \ref{lem_bdd}}
    First, we show that the optimal cost-to-go value is bounded by $\frac{1}{1-\gamma}$. For any fixed policy $\pi$, we have 
    $$
        V^{\pi}(s) 
        =
        \mathbb{E}
        \left[
            \sum\limits_{t=0}^{\infty} \gamma^{t}R(s_t,a_t,s_{t+1})
            \vert s_0 = s
        \right]
        \leq
        \frac{1}{1-\gamma},
    $$
    where the last inequality is obtained by the assumption that the reward function is bounded by 1. The inequality also holds for the optimal policy, which implies $\|\bm{V}^*\|_{\infty}\leq\frac{1}{1-\gamma}$.
    
    Then, we show that the estimated cost-to-go value obtained by our algorithm is bounded by $\frac{1}{1-\gamma}$ by contradiction. Let $\tau$ be the smallest index such that 
    \begin{align}
        \label{stcond1}
        \|\bm{V}_{\tau}\|_{\infty}>\frac{1}{1-\gamma}    
    \end{align}
    if \(\tau\) indexes a global iteration, or
    \begin{align}
        \label{stcond2}
        \|\bm{W}_{\tau}\|_{\infty}>\frac{1}{1-\gamma}
    \end{align}
    if it indexes an aggregated iteration. If the $\tau$-th iteration is a global iteration, we have $\|\bm{V}_{\tau-1}\|_{\infty}\leq\frac{1}{1-\gamma}$ and so
    \begin{align*}
        |V_{\tau}(s)|
        &=
        |\min_{a\in\mathcal{A}} R(s,a) + \gamma\bm{P}_{s,a}^{\top}\bm{V}_{\tau-1}|\\
        &\leq
        \max_{a\in\mathcal{A}}\left\{|R(s,a)| 
            + |\gamma\bm{P}_{s,a}^{\top}\bm{V}_{\tau-1}|
            \right\}\\
        &\leq
        1+\gamma\max_{a\in\mathcal{A}}\|\bm{P}_{s,a}\|_1\|\bm{V}_{\tau-1}\|_{\infty}\\
        &\leq
        1+\frac{\gamma}{1-\gamma}\\
        &=
        \frac{1}{1-\gamma},
    \end{align*}
    for all $s\in\mathcal{S}$. Here, the first line is obtained by the definition of value iteration process; the second line is obtained by the triangle inequality; the third line holds because $r(s,a) \leq 1$ and because Holder's inequality gives that $|\bm{P}_a\bm{V}_{\tau-1}|\leq\|\bm{P}_{s,a}\|_1\|\bm{V}_{\tau-1}\|_{\infty}$ for all $s\in\mathcal{S}$ and $a\in\mathcal{A}$; the fourth line follows from $\|\bm{P}_{s,a}\|_1=1$ and the condition that $\|\bm{V}_{\tau-1}\|_{\infty}\leq\frac{1}{1-\gamma}$; and the last line is obtained by direct calculation. However, this inequality contradicts Eq.~\eqref{stcond1}.
    
    Alternatively, if the $\tau$-th iteration is a aggregated iteration, we find the contradiction in two cases. On the one hand, if the $\tau$-th iteration is the first iteration in a aggregated process or $\tau=\min\{\mathcal{A}_i\}$ for some $i$, we have $\|\bm{V}_{\tau-1}\|_{\infty}\leq\frac{1}{1-\gamma}$. Then, for any aggregated state $j$, from Algorithm 2 we have 
    \begin{align*}
        |\tilde{\bm{V}}(\bm{W}(\bm{V}_{\tau - 1}))(s)|
        &\leq
        \|\bm{V}_{\tau-1}\|_{\infty} - \frac{\varepsilon}{2}
        \leq
        \frac{1}{1-\gamma},
    \end{align*}
    which also contradicts Eq.~\eqref{stcond2}. On the other hand, if $\tau$ is such that the $\tau-1$-th iteration is also a aggregated iteration, we can use a similar method as in the global iteration case to derive a contradiction. Thus, we have that the estimated cost-to-go value obtained by our algorithm is bounded by $\frac{1}{1-\gamma}$.

\subsection{Proof of Lemma \ref{lem}}
The proof of lemma \ref{lem} follows an induction argument on a similar inequality: For all $n \geq 1$, we have 
\begin{align}\label{ineq_induction}
    ||\bm{V}_{t_{n+1}} - \bm{V}^*||_{\infty} \leq \gamma^{n}||\bm{V}_{\tau_1} - \bm{V}^*||_{\infty} + \sum_{i=1}^{n-1} \gamma^i \varepsilon.
\end{align}
For the initial case, we have that 
$$
    ||\bm{V}_{t_{2}} - \bm{V}^*||_{\infty} \leq \gamma||\bm{V}_{\tau_{2}} - \bm{V}^*||_{\infty} \leq \gamma(||\bm{V}_{t_1} - \bm{V}^*||_{\infty} + \varepsilon) \leq \gamma^2||\bm{V}_{\tau_1} - \bm{V}^*||_{\infty} + \gamma\varepsilon.
$$
For $n > 1$, by assuming $\eqref{ineq_induction}$, we have that

\begin{equation*}
    \begin{aligned}
    ||\bm{V}_{t_{n+2}} - \bm{V}^*||_{\infty} &\leq \gamma||\bm{V}_{\tau_{n+2}-1} - \bm{V}^*||_{\infty} \leq 
    \gamma(||\bm{V}_{t_{n+1}} - \bm{V}^*||_{\infty} + \varepsilon)\\
    &\leq \gamma^{n+1}||\bm{V}_{\tau_1} - \bm{V}^*||_{\infty} + \sum_{i=1}^{n} \gamma^i \varepsilon,
    \end{aligned}
\end{equation*}
where the second last inequality comes from Eq.~\eqref{ineq_recur}.
After showing Eq.~\eqref{ineq_induction}, the lemma follows upon noticing that $\sum_{i=1}^{n} \gamma^i \varepsilon < \frac{\varepsilon}{1-\gamma}$.

% \begin{figure}[t]
% \centering
%   \includegraphics[width=.3\linewidth]{cartpole.png}
%   \caption{Cartpole Problem. The goal is to balance the pendulum system with the pole with pendulum attached starting upright at the beginning of the problem.}
%   \label{fig:cart}
% \end{figure}

\end{document}